%% file: cvmpaper_finalcopy.tex
\documentclass[10pt,twocolumn,letterpaper]{article}

\usepackage{cvm}
\usepackage{times}
\usepackage{epsfig}
\usepackage{graphicx}
\usepackage{amsmath}
\usepackage{amssymb}
\usepackage{booktabs}
\usepackage{multirow}
\usepackage{adjustbox}
\usepackage{makecell} 
\usepackage{algorithm}
\usepackage{algorithmic}
\usepackage{rotating}

\usepackage[pagebackref=true,breaklinks=true,letterpaper=true,colorlinks,bookmarks=false]{hyperref}

\cvmfinalcopy

\def\cvmPaperID{****} 

\ifcvmfinal\fi
\begin{document}
\newcommand{\sfm}{\mathrm{SFM}}
\title{Active Self-Training for Weakly Supervised 3D Scene Semantic Segmentation}

\author{Gengxin Liu\\
Shenzhen University\\
Shenzhen, China\\
{\tt\small gengxin.v.liu@gmail.com}
\and
Oliver van Kaick\\
School of Computer Science Carleton University\\
Ottawa, Canada\\
{\tt\small Oliver.vanKaick@carleton.ca}
\and
Hui Huang\\
Shenzhen University\\
Shenzhen, China\\
{\tt\small hhzhiyan@gmail.com}
\and
Ruizhen Hu\\
Shenzhen University\\
Shenzhen, China\\
{\tt\small ruizhen.hu@gmail.com}
}

\maketitle

\begin{abstract}
   	Since the preparation of labeled data for training semantic segmentation networks of point clouds is a time-consuming process, weakly supervised approaches have been introduced to learn from only a small fraction of data. These methods are typically based on learning with contrastive losses while automatically deriving per-point pseudo-labels from a sparse set of user-annotated labels. In this paper, our key observation is that the selection of what samples to annotate is as important as how these samples are used for training. Thus, we introduce a method for weakly supervised segmentation of 3D scenes that combines self-training with active learning. The active learning selects points for annotation that likely result in performance improvements to the trained model, while the self-training makes efficient use of the user-provided labels for learning the model. We demonstrate that our approach leads to an effective method that provides improvements in scene segmentation over previous works and baselines, while requiring only a small number of user annotations.
\end{abstract}

\section{Introduction}

In recent years, we have seen the introduction of approaches for semantic segmentation of point clouds, which have been quite successful in providing meaningful segmentations of indoor scenes~\cite{qi17pp,li2018,thomas2019,han2020}. Much of this success is due to the use of deep learning methods combined with the availability of large amounts of labeled data, e.g., datasets such as ScanNet~\cite{dai2017scannet} and S3DIS~\cite{armeni2017joint}. However, the applicability and scalability of these methods to new contexts is limited, since creating training data is a time-consuming task, involving the manual labeling of points.

To address the dependency of segmentation methods on large amounts of training data, methods for weakly supervised segmentation have been introduced, which require only a fraction of the training data commonly used. These methods either estimate pseudo-labels for the data in order to train segmentation networks~\cite{wei2020,xu2020,gadelha2020}, or use variations of a contrastive loss for enabling learning transfer~\cite{xie2020,jiang2021,hou2021exploring}. 

\input{figures/teaser}

The recent ``one thing one click'' method~\cite{liu2021one} introduces an iterative self-training approach that alternates between training and label propagation, where the labels from points annotated by the user are propagated to the unlabeled data based on a learned data similarity measure. The method achieves some of the best results among weakly supervised methods by training on data with only one label per object. However, the user is responsible for manually selecting one point per object, which can be difficult in cluttered scenes. 
Moreover, the method is quite complex, involving the combination of two networks, one for semantic segmentation and one for similarity learning, trained in an iterative manner with a contrastive loss. The extra complexity of training a similarity estimation network is necessary for enabling an accurate label propagation during the self-training.

In this paper, we introduce a method for weakly supervised segmentation that combines self-training~\cite{liu2021one} with active learning~\cite{hu18semantic,wu2021,shi2021label}. Our focus is on improving the selection of samples to be annotated while simplifying the label propagation step. The active learning method that we introduce automatically selects the points that have to be annotated by the user according to an uncertainty measure, reducing the amount of work involved in the annotation task, and querying the user for annotations of the points that likely lead to considerable performance improvements in the model. In addition, our self-training method is much simpler, requiring only the training of a segmentation network. We perform the label propagation based on the geometry of the scenes via super-voxels, without the need to train a similarity estimation network, which is unnecessary for the final segmentation task. 

We demonstrate that the active learning combined with our simpler self-training pipeline leads to improved point cloud segmentation results of indoor scenes,
when compared to previous approaches and baselines. As summarized in Fig.~\ref{fig:teaser}, we obtain higher mIoUs than previous works on two established datasets of indoor scenes~\cite{dai2017scannet,armeni2017joint}.
In addition, we show that selecting points according to the active learning, which can potentially query multiple points for the most challenging objects to segment, leads to improved results compared to selecting a single point per object, while still requiring the same or smaller total number of user annotations.

\section{Related work}

\noindent\textbf{Understanding 3D scenes.} 3D scenes are commonly scanned from the environment and represented as point clouds, and their understanding involves solving problems such as 3D object detection, classification, semantic segmentation, and instance segmentation. 
The problem mostly related to our paper is semantic segmentation of point clouds, and thus we discuss it in more detail here. 

Earlier solutions for scene segmentation transform point clouds into volumetric grid representations for processing with convolutional neural networks (CNNs)~\cite{wu2015}. However, since it is unnatural to represent sets of points as a volumetric grid, several approaches were introduced for directly processing point clouds, such as variations of PointNet~\cite{qi17,qi17pp} that make use of the symmetric max pooling operation, or generalizations of convolution to sets of points, such as PointCNN~\cite{li2018}, PointConv~\cite{wu2019}, KPConv~\cite{thomas2019}, A-CNN~\cite{komarichev2019}, SPLATNet~\cite{su2018}, and hybrid approaches such as Point-Voxel CNN~\cite{liu2019} and OccuSeg~\cite{han2020}. 3D-RNN~\cite{ye20183drnn} uses two-direction hierarchical recurrent neural networks to extract long-range spatial dependencies in the point cloud. 
PCT~\cite{guo2021pct} designs a point cloud processing architecture comprised of Transformers. Peng et al.~\cite{peng2020semantic} proposes a part-level semantic segmentation annotation method for single-view point cloud using the guidance of labeled synthetic models.


Finally, a few methods are also based on other data representations, such as multiple 2D views~\cite{kundu2020}, combined 2D/3D information~\cite{dai2018}, hash tables that enable sparse convolutions~\cite{graham2018}, and graphs~\cite{landrieu2018}. Tatarchenko et al.~\cite{tatarchenko2018} project features into predefined regular domains and apply 2D CNNs to the domain.
Some works focus on online segmentation, which aims to perform real-time 3D scene reconstruction along with semantic segmentation~\cite{huang2021supervoxel, zhang2020fusion}.
We base the backbone of our segmentation network on the 3D U-Net architecture~\cite{choy20194d}, which implements the efficient \textit{generalized sparse convolution}.

\noindent\textbf{Weakly supervised segmentation.} 
In our paper, we address the weakly-supervised semantic segmentation of point clouds with limited annotations. Recently, a few methods have been introduced that made significant advancements towards solving this problem. 
Xu et al.~\cite{xu2020} directly label a small amount of points (around 10\% of the data) and train an incomplete supervision network with spatial smoothness constraints, showing that the learning gradient of the insufficient supervision approximates the gradient of the full supervision.
Other methods generate pseudo-labels for the unlabeled data based on a small set of labeled samples. For example, Wei et al.~\cite{wei2020} perform weak supervision with labels that only indicates what classes appear in the training samples. These labels are transformed into point pseudo-labels with a region localization method.
Cheng et al.~\cite{cheng2021sspc} propagate sparse point labels to the unlabeled data using a graph of superpoints extracted from the point cloud.

\input{figures/overview}

Another important line of work investigates the use of {\em contrastive losses} for learning transfer when processing point clouds. Xie et al.~\cite{xie2020} performed the first studies in this regard, showing that learning transfer is possible for point cloud processing. Moreover, Jiang et al.~\cite{jiang2021} use a contrastive loss to guide semantic segmentation, while Hou et al.~\cite{hou2021exploring} extend the contrastive loss to integrate spatial information. Zhang et al.~\cite{zhang2021} use a self-supervised method in a contrastive framework to pre-train point cloud processing networks with single-view depth scans. The networks can then be fine-tuned for multiple tasks such as segmentation and object detection.
Liu et al.~\cite{liu2021one} combine self-training based on contrastive learning with a label-propagation mechanism in the ``one thing one click'' method, achieving some of the best results on point cloud segmentation. In contrast, we introduce a method that provides a more effective selection of labeled samples with active learning, while simplifying the label propagation, achieving significant improvements over previous works.

\noindent\textbf{Active learning.} Active learning approaches seek to minimize manual annotation efforts by strategically querying the user for the annotations that maximize the performance of the learned models. This is typically an iterative process involving the selection of what points have to be labeled and then updating a model based on the new annotations.
Earlier active learning approaches for segmentation focused on the annotation of 3D shapes, such as the method of Yi et al.~\cite{yi2016} that queries users for annotation and verification. More recent approaches focus on the annotation of point clouds of scanned objects, e.g., as the method of Hu et al.~\cite{hu18semantic} that simultaneously performs reconstruction and segmentation. 

For point clouds of entire scenes, Wu et al.~\cite{wu2021} introduced an active learning approach that measures the uncertainty in the point cloud labeling and selects diverse points to minimize redundancy in the point selection. Shi et al.~\cite{shi2021label} maximize model performance for a limited annotation budget by measuring consistency at the super-point level. We also incorporate iterative active learning into our method, although we measure uncertainty based on the stochastic behavior of a segmentation network, leading to an effective sample selection for segmentation.



\section{Self-training with active learning}
\label{sec:method}


An overview of our method is illustrated in Fig.~\ref{fig:overview}. Our goal is to train a 3D semantic segmentation network (3D U-Net in Fig.~\ref{fig:overview}) that is able to predict accurate semantic labels for the input point cloud $X$ representing a 3D scene. Our key insight is that, when training the segmentation network with extremely sparse labels, 
the selected annotation labels have a substantial impact in the accuracy of the segmentation model.
Thus, the segmentation network is trained in an iterative manner with self-training combined with active learning, according to the following steps. 

As a pre-processing step, given the input point cloud $X$, as shown in Fig.~\ref{fig:overview} (a), we first over-segment $X$ into geometrically homogeneous super-voxels. 
The super-voxels will be used for label propagation, where all the points located in the same super-voxel will share the label provided by the user for a point in the super-voxel. Note that the active learning ensures that the user will never be required to annotate more than one point in a super-voxel.

During each iteration of the self-training, the segmentation network is trained
with two sets of per-point labels: $T$ and $P$, as in Fig.~\ref{fig:overview} (f) and (i).
The set of true labels $T$ is obtained from user annotations of a sparse set of sample points $\hat{T}$.
Based on the uncertainty predicted by the network trained in the previous iteration, we  sample a set of points for user annotation.

The true labels $\hat{T}$ of these samples are then propagated within their containing super-voxels to yield a set of propagated per-point labels $T$, as shown in Fig.~\ref{fig:overview} (d)-(f). 
The set of pseudo labels $P$ is taken from the prediction results of the segmentation network trained in the previous iteration of the self-training, where labels are selected based on the prediction confidence, and is empty in the first iteration. 

We repeat the iterations composed of user annotation and network
training until reaching a pre-defined number of iterations,
to satisfy a requested annotation budget.
We provide more details on the components of the method as follows.



\subsection{3D semantic segmentation network}

We adopt the 3D U-Net architecture~\cite{choy20194d} as the backbone of our segmentation network. 
The input to U-Net is a point cloud $X$ of $N$ points, with each point  $x_i$ containing both 3D coordinates and color  information, where $i \in \{1, ..., N \}$. 
The network predicts the probability of each semantic category for each point $x_i$, denoted as $p_{i, c}$, and the probability corresponding to the ground truth category $\bar{c}$, provided either by $T$ or $P$, is denoted as $p_{i, \bar{c}}$. The network is then trained with the softmax cross-entropy loss: 
\begin{equation}
L = \frac{1}{\vert T\vert}\sum_{i\in T}-\log{p_{i,\bar{c}}} +\lambda \frac{1}{\vert P\vert}\sum_{i\in P}-\log{p_{i,\bar{c}}},
\end{equation}
where $\lambda$ is a combination weight.

In the first iteration of the self-training, the network is trained with the set $T$ derived from a set of randomly sampled points $\hat{T}$ annotated by the user, and the set $P$ is empty. In the subsequent iterations, the set $\hat{T}$ is expanded with new samples annotated by users, where the samples are selected via active learning as explained in Section~\ref{sec:active}, which are then propagated through the super-voxels to form a new set $T$.
The set $P$ is updated with the prediction results of the current network, as explained in Section~\ref{sec:self}.

Note that, during inference, the average of predictions of all the points inside the same super-voxel is used as the prediction result of the super-voxel, so that all the points in the same super-voxel have the same prediction. 
This ``voting method'' ensures that the prediction for the points inside each super-voxel is consistent, which improves the final prediction accuracy, as we show in our ablation studies.


\subsection{Active learning for true label annotation}
\label{sec:active}

During the active learning process, the user is asked to annotate a sparse set of points with labels, illustrated in Fig.~\ref{fig:overview} (d).
To select the most effective set of points to annotate for improving the accuracy of the segmentation, we measure the uncertainty of the labeling of each point based on the current prediction results.
The uncertainty of each point is measured by calculating the standard deviation of several stochastic forward passes, and using the one corresponding to the category  with the highest mean prediction confidence. 
More specifically, for each input point cloud, we first get $K$ different versions via standard data augmentation operations~\cite{choy20194d}, and then for each point, we compute the mean and standard deviation of these $K$ probability distributions predicted from those $K$ different input versions.
Finally, the standard deviation of the category with the highest mean probability is used as the uncertainty $u_i$ of the point $x_i$:
\begin{equation}
u_i = \sqrt{\frac{ \sum_{k}{( p_{i, \hat{c}}^{k} - \overline{p}_{i,\hat{c}}  )^2  } }{ K }},
\label{eq:uncertainty}
\end{equation}
where $p_{i, c}^{k}$ is the predicted probability
of point $x_i$ in the $k$-th point cloud version for category $c$ and $\overline{p}_{i, c} = \frac{\sum_k p_{i, c}^{k}}{K} $ is the mean 
probability for the $K$ versions for category $c$, with $\hat{c}$ being the category with highest mean probability.

For each iteration, we select $m$ points according to the uncertainty distribution over points, since intuitively points with high uncertainty require more reliable user input.

\subsection{Pseudo label generation}
\label{sec:self}

Similar to~\cite{liu2021one}, we iteratively update the set of pseudo labels $P$.
Starting with the label predictions of the segmentation network trained at a given iteration, we take the predictions with high confidence (larger than a given threshold $\tau$) and use them as updated pseudo labels $P$. The pseudo labels $P$ are then used together with the labels $T$, propagated from the true labels $\hat{T}$, to train the network in the next iteration. Note that we also use the mean prediction probability $\overline{p}_{i,\hat{c}}$ of $K$ different point cloud versions, as in Eq.(\ref{eq:uncertainty}), to compute the confidence in this step.

To limit error propagation in our iterative training and pseudo-labeling process, we generate new labels for all unlabeled samples and reinitialize the neural network after each pseudo-labeling step, similarly to Rizve et al.~\cite{rizve2021defense}.

\section{Results and evaluation}

\noindent\textbf{Datasets.} 
Our experiments are conducted on ScanNet-v2~\cite{dai2017scannet} and S3DIS~\cite{armeni2017joint}, which allows us to compare our results directly to Liu et al.~\cite{liu2021one} and other methods.
We use the original training-validate-test split provided in these two datasets. 
One thing to note is that we focus on the ``Data Efficient'' annotation setting as in~\cite{hou2021exploring}, which is a more realistic setting than the ``one thing one click'' setting in~\cite{liu2021one}, since~\cite{liu2021one} requires the user to identify each individual object in the scene.
Regarding the super-voxel creation, for the ScanNet dataset, we use the method of~\cite{dai2017scannet}, while for S3DIS, we use the method of Landrieu et al.~\cite{landrieu2018}, as in~\cite{liu2021one}. 

\noindent\textbf{Implementation details.} 
We implement our method with the PyTorch \cite{paszke2019pytorch} framework based on the implementation of Choy et al.~\cite{choy20194d}. We use the default data augmentation of~\cite{choy20194d} and set the batch size to be $4$ for both ScanNet-v2 and S3DIS datasets. The number of training iterations on Scannet-v2 and S3DIS are 6 and 5, respectively.
For ScanNet-v2, the initial learning rate is 0.1, with polynomial decay with power 0.9, and the model is trained for a total of 100K steps in each iteration. 
For S3DIS, the initial learning rate is 0.03, with polynomial decay with power 0.9, and the model is trained for a total of 60K steps in each iteration. 
Uncertainty and confidence are computed from $K=5$ different versions’ predictions. The threshold $\tau$ for pseudo label generation is set to be $0.99$ in the first few iterations and $0.95$ in the last two iterations, to generate more training data with the refined segmentation network.
The combination weight $\lambda$ is set to be $0.5$.

In the following sections, we first report the comparison results with existing methods in Section~\ref{sec:compare} and conduct ablation studies in Section~\ref{sec:ablation} on both datasets. Then, in Section~\ref{sec:one}, we show that our method can also work in the ``one thing one click'' setting and obtain better results.

\subsection{Comparison with existing methods }
\label{sec:compare}

Note that, for a fair comparison, we use less or the same amount of user annotations as the existing methods. The actual amount of user annotations used is reported in all the tables. 
As our method uses active learning to select samples to annotate, the number of samples is evenly distributed. 
For example, if $n$ is the total amount of user annotations and $k$ is the number of iterations, then $m = n/k$ points are sampled in each iteration for users to annotate. 

\input{figures/tab_comp_scannet}

\input{figures/comp}

\noindent\textbf{Results on ScanNet-v2.} 
Table~\ref{tab:comp_scannet} reports the benchmark results on the ScanNet-v2 test set,
where the existing methods are roughly divided into two groups: (i) Fully supervised approaches with 100\% supervision, including the state-of-the-art networks for point cloud segmentation, and (ii) Weakly supervised approaches, including the most recent work 1T1C~\cite{liu2021one} and methods using contrastive pre-training followed by fine-tuning with limited labels~\cite{hou2021exploring,xie2020}. 
Note that ``our fully-supervised baseline'' refers to the segmentation network~\cite{choy20194d} which we take as the backbone of our method, trained with 100\% supervision.

Our method produces very competitive results with only 20 labeled points per scene. Firstly, our method outperforms the best weakly supervised approaches on the ``Data Efficient'' setting with 20 points by nearly 11\% mIoU (70.3\% vs.\ 59.4\%). Our method also surpasses 1T1C by nearly 1\% mIoU (70.3\% vs.\ 69.1\%) when 1T1C uses twice the amount of user annotations, i.e., 0.02\% vs.\ 0.01\% (20 points/scene), and requires object instance information. Secondly, the gap between our method and full supervision is less than 3.3\% mIoU (70.3\% vs.\ 73.6\%), which shows the effectiveness of our method.


Fig.~\ref{fig:comp} shows visual examples of results obtained with our method and the fully-supervised baseline, compared to the ground truth. We can see that our method obtains comparable, sometimes even better, results than the fully-supervised baseline.
For example, for the scene shown in the first row, our method is able to correctly label all the chairs in different arrangements. 
Our method is also able to recognize the cabinet for the scene shown in the second row, although the fully-supervised baseline misclassifies it as a counter.


\input{figures/tab_comp_S3DIS}

\noindent\textbf{Results on S3DIS.} 
Table~\ref{tab:comp_S3DIS} reports the comparison with existing methods on the S3DIS dataset.
We also compare with fully supervised approaches and weakly supervised approaches. 
For weakly supervised approaches, we only compare with 1T1C~\cite{liu2021one}, since it's the most recent work and performs better than previous approaches.

We can see that with only 20 labeled points per scene (0.01\% supervision), our method outperforms 1T1C (0.02\% supervision) by nearly 6\% mIoU. It also surpasses the 1T3C instance that annotates 3 random points per object and results in 0.06\% supervision. 
In addition, our approach even outperforms several fully-supervised methods, which again demonstrates the effectiveness of our method.



Fig.~\ref{fig:comp} shows visual results of our method on the S3DIS dataset. 
We see that even with more challenging and crowded scenes, our method can still obtain reasonably good prediction results similar to those obtained via the fully-supervised baseline.


\subsection{Ablation studies}
\label{sec:ablation}
Our ablation studies are conducted on the most challenging setting with only 20 points annotated in each scene, for both the ScanNet-v2 and S3DIS datasets.
For ScanNet-v2, the evaluation is conducted on the validation set.
For S3DIS, the evaluation is conducted on Area 5.

We test the two key components of our method: the self-training (denoted ``Self-train.'') and the active learning (``Active learn.''). We also test the ``voting'' process used during the inference.  
The results are shown in Table~\ref{tab:ablation}. We see that each component of our method contributes to the quality of the final results.

\input{figures/tab_ablation}
\input{figures/sample}

More specifically, in the first row, we show the results of our baseline method, which is the segmentation network~\cite{choy20194d} trained with all the user annotations, where the samples are either selected with a pre-trained model provided by~\cite{hou2021exploring} for ScanNet-v2 or randomly selected for S3DIS.
By adding ``voting'' of predictions within each super-voxels, we can see that the results improve, which confirms that maintaining label consistency inside of each super-voxel contributes to an improvement of the results.
When adding one of our two key components individually in the third and forth row, we see that the performance is improved on both datasets in either case, while the performance is the best when using the full method with both key components.

%

\noindent\textbf{Discussion on sample selection.} 
Here we investigate more the sample selection via active learning, to provide insights on why our method outperforms previous methods.

Fig.~\ref{fig:sample} shows statistics on the classes of objects where the points were selected with the active learning in our method, compared to the method of Hou et al.~\cite{hou2021exploring}.
We see that our method selects less samples on floors while selecting more samples on categories like wall, door, window, desk, counter, and sink, which leads to significant improvement of prediction accuracy for these categories. Our method even improves the prediction accuracy for floors as this class often gets misclassified as ``door'' in early-iteration results, while in the final results, there is less confusion with doors as our method selects more samples for doors.

We also show a visual comparison of selected samples in Fig.~\ref{fig:sample}. 
For our method, we show the uncertainty map over the scene, which is used to guide the sample selection, together with the selected samples for each iteration.
Comparing to the samples selected at once with the pre-trained model in~\cite{hou2021exploring}, we see that the samples that we select are located on more  complicated objects with more visual variability like chairs rather than the floor.

Note that, if we have $s$ scenes in the training set, there are two ways for active learning to sample the points. One way is to sample the same number of points in each scene, i.e., $n/(k*s)$ per scene, and the other way is to sample the points among all the scenes based on uncertainty only, which results in different numbers of sampled points in different scenes. We tested these two different options and found that the results are similar. 

This is because both datasets have large scene variations, which leads to samples evenly distributed over the scenes even if they are sampled over the entire dataset based on pointwise uncertainty.
If another dataset with scenes similar to each other were given, we believe that the results with sampling over all the scenes would be better. 


\subsection{Results under ``one thing one click'' setting}
\label{sec:one}

In this section, we test our method under the ``one thing one click'' setting. The only change in our method is that, for the sample selection during the active learning, we avoid objects that have been sampled before, selecting the sample with highest uncertainty among the remaining objects. In other words, our active learning chooses which object to sample as well as which point inside each object should be sampled in the each iteration. 

\input{figures/tab_1T1C}

In more detail, we first compute the average number $n_o$ of objects per scene for both datasets, which results in $n_o = 32$ for ScanNet-v2 and $n_o = 36$ for S3DIS.
Here we set the number of iterations to be $k = 6$. For each of the first five iterations, we select $6$ samples from $6$ different objects in the scene and set them to be invalid for the selection in the next iteration. In the final iteration, we select one point with highest uncertainty from each remaining object to complete the ``one thing one click'' sample selection.

\input{figures/sample_within_object}

Table~\ref{tab:1T1C} shows how the performance changes across the iterations. 
We see that, for the S3DIS dataset, our method already obtains better results than 1T1C~\cite{liu2021one} when only $24$ points were annotated in iteration 4. 
Fig.~\ref{fig:sample_within_object} shows a visual comparison of a sample point selected by our method and the point randomly sampled as in~\cite{liu2021one}.
We can see that our method learns to select points that are located in more important regions inside the objects, which is more informative and leads to more accurate prediction results.
By focusing on labeling more challenging regions, our method can correctly predict other unlabeled regions, while with random samples as in~\cite{liu2021one}, the method may not be able to extract enough information to correctly predict the labels for the challenging regions.

One interesting result that we observed is that, for both the ScanNet-v2 and S3DIS datasets, after adding $24$ points while constraining only one point per object, the results are worse than for our method under the 20 points/scene setting, where less annotations are given. 
To find the reason behind this behavior of the method, in Fig.~\ref{fig:sample_number}, we show the statistics of number of points sampled on each object for our method under the 20 points/scene setting.
We see in the results for ScanNet-v2 that, compared to the ``one thing one click'' setting where $100\%$ of the objects get exactly one click, most of the objects ($61\%$) in our setting do not get any samples, $14\%$ of the objects get more than one click, and only $25\%$ of the objects get exactly one click. We observe similar distributions on S3DIS. 
This indicates that the ``one thing one click'' procedure may not be a good way to sample points for semantic segmentation. 
Complicated objects may need more sample points than others, while for objects coming from simpler categories, we do not require annotations for each scene.

\input{figures/sample_number}


\section{Conclusion}

We introduced a weakly supervised method for semantic segmentation of 3D scenes, which is the combination of an active learning component that selects the most effective points to be annotated by users, and a self-training approach that makes efficient use of the user labels. We showed that our method leads to improvements of 11\% and 6\% mIoU over previous works on well-known datasets, while using the same amount or less of user annotations. Our method is also competitive with fully supervised methods. 


\noindent\textbf{Limitations and future work.}
Currently, in each iteration,
the self-training uses the labels collected from the active learning and the pseudo labels only once. However, it is possible to use the active learning as an outer loop and self-training as an inner loop of the method, to run more iterations of self-training for each set of annotated points. This may lead to better results at the expense of increasing the training time.
It is also possible to use a pre-trained model as in~\cite{hou2021exploring} to obtain a better set of initial samples for annotation.
For time efficiency, theoretically, the work of~\cite{liu2021one} requires extra time to identify each individual object when annotating the same number of points as in our method, while the drawback of our point selection based on active learning is that the points need to be annotated in several iterations instead of once as in~\cite{liu2021one}. 
Although we believe that the annotations can be collected through crowdsourcing on the internet and thus no extra user effort is needed as users don't have to wait in front of the computer,
this would cause information delay and would become a problem when the collection is conducted in person. 
Moreover, we adopt the same supervoxel clustering methods as in~\cite{liu2021one} for a fair comparison, however, it's worth exploring other more advanced methods such as~\cite{huang2021supervoxel} and ~\cite{lin2018toward} to further boost the performance.


\appendix

\subsection*{Appendix}
In this supplementary document, we first present more details of our ActiveST framework in Section~\ref{sec:detail_activest}. Then, we report detailed benchmark results on the ScanNet-v2 test set with per-category performance of our method comparing to other weakly supervised methods on the most challenging setting with only 20 points annotated in each scene in Section~\ref{sec:per_category}. Finally, we show more results on both ScanNet-v2 and S3DIS datasets with various numbers of annotated points in Section~\ref{sec:scannet_s3dis}. 

\section{Details of our ActiveST framework}
\label{sec:detail_activest}

We present the training procedure for our proposed ActiveST framework in Algorithm~\ref{algorithm:activest}. Fig.~\ref{fig:network} shows the 3D U-Net architecture we used as the backbone. This architecture was proposed in~\cite{choy20194d} for semantic segmentation, which contains four blocks for encoding and four blocks for decoding.  For each block, we show the output dimension $D$ and number of convolution layers $N$. 
\input{figures/network}
\input{figures/algorithm}
\input{figures/tab_sup_percategory}

\section{Per-category result on ScanNet-v2}
\label{sec:per_category}
In this section, we present detailed per-category performance of our method comparing to other weakly supervised methods on the ``Data Efficient'' setting with 20 labeled points per scene as supplement.

The results are shown in Table~\ref{tab:sup_percategory}. We can see that our method gets significant performance improvement on categories such as ``bathtub'',  ``counter'',  ``cabinet'',  ``sink'' and  ``window''.
We find that the reason is that our method is able to sample more points on those more challenging categories via  active learning, when comparing to the default set of sampled points used in other methods. 

We also notice that ``picture'' is the only set that our method gets worse performance than the method of 1T1C~\cite{liu2021one}.  
We find that this is because our method puts more samples on other categories that lead to more confusion to the output prediction, which leads to smaller number of annotations on the ``picture'' category. But as we will shown in the following section, when given more annotation budget, more points on the ``picture'' category are sampled and thus the results are highly improved.

\input{figures/sup_comp_active_benchmark}


\section{More results on ScanNet-v2 and S3DIS }
\label{sec:scannet_s3dis}
In this section, we show the results of our method on both ScanNet-v2 and S3DIS datasets with increasing number of annotated points, including 20, 50, 100 and 200 points/scene. 


\input{figures/tab_sup_scannet}

\input{figures/tab_sup_s3dis}

We first report the results on the ScanNet-v2 test set. 
As shown in Table~\ref{tab:sup_scannet}, with increasing number of sampled points, the performance is consistently improved for all the method, while our method always gets highest mIoU under all settings. 
We can see that our method surpasses all other weakly supervised approaches in the same setting by a large margin and set a new SOTA in the Scannet-v2 ``Data-efficient'' challenge. 
Note that, when only using 20 labeled points per scene, our method even beats the most competitive weakly supervised approach 1T1C~\cite{liu2021one} which it uses 10 times points (200 labeled points per scene) by nearly 1\% mIoU (70.3 vs. 69.4\%).

With more annotation budget, we observe that our method tends to select more points on ``picture'', which is the category that gets worst perform under the ``20 pts'' setting as discussed in~\ref{sec:per_category},  and then mIoU of ``picture'' gains 25.6\% improvement (12.6\% vs. 38.2\%) from ``20 pts'' to ``200 pts''. 
Other than the ``picture'' category, the percentage of points sampled on the ``window'' category also increases, and  mIoU of ``window'' gets improved by 5.6\% (64.5\% vs. 70.1\%) from ``20 pts'' to ``200 pts''. 
Fig.~\ref{fig:sup_comp_active_benchmark} shows the comparison of samples selected by our method and~\cite{hou2021exploring} under different settings. It can be observed that our method selects more samples on ``picture'' and ``window'' instead of ``floor'' under each setting, thus the performance of these categories gains great improvement in the final results.

We also notice that although the sample proportion of some categories such as ``toilet'',``cabinet'' and ``curtain'' become smaller, the performance over those categories is not sacrificed as our model is less likely to misclassify objects in those categories as ``wall'' as before due to more samples on ``wall'' category. 
We also report our results on S3DIS under those various settings, as shown in Table~\ref{tab:sup_s3dis}. With more points selected by our method, the gap between our method and full-sup baseline is further reduced.

\subsection*{Acknowledgements}
We thank the anonymous reviewers for their valuable comments. This work was supported by NSFC (61872250, U2001206, U21B2023), GD Natural Science Foundation (2021B1515020085), DEGP Key Project (2018KZDXM058, 2020SFKC059), Shenzhen Science and Technology Innovation Program (JCYJ20210324120213036), the Natural Sciences and Engineering Research Council of Canada (NSERC), and Guangdong Laboratory of Artificial Intelligence and Digital Economy (SZ).

\subsection*{Declaration of competing interest}

The authors have no competing interests to declare that are relevant to the
content of this article.\\


%
%

\end{document}

%% file: figures/teaser.tex
\begin{figure}[!t]
	\centering
	\includegraphics[width=\linewidth]{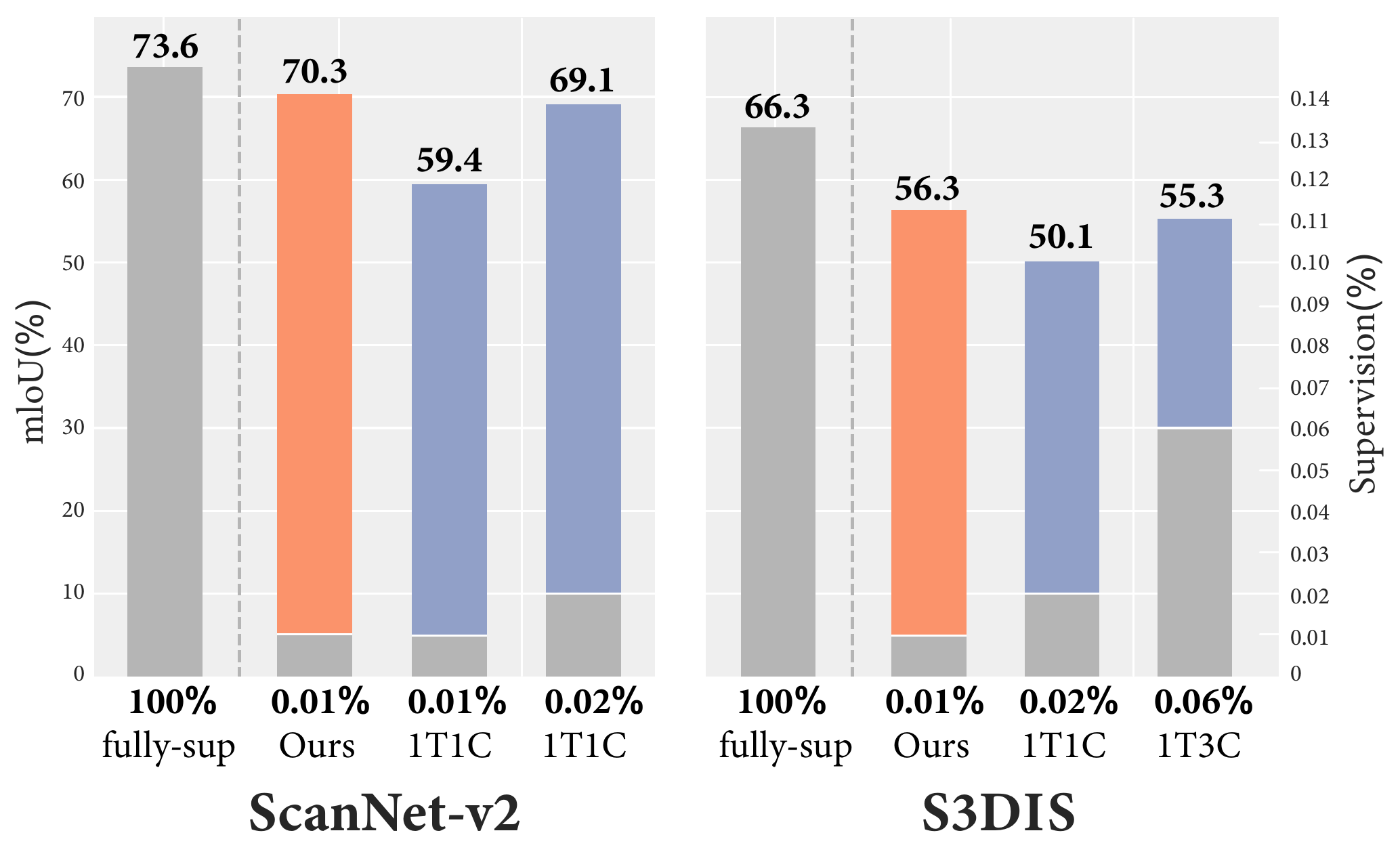}
        \caption{Comparison of our weakly supervised semantic segmentation method to the recent ``one thing one click'' method~\cite{liu2021one} (denoted 1T1C or 1T3C depending on the amount of annotation data used) and a fully-supervised method with the same backbone as ours~\cite{choy20194d}, on two datasets of 3D scenes. Our method achieves better results (left axis) while using an equal or smaller number of user annotations than other weakly supervised methods (right axis). }
	\label{fig:teaser}
\end{figure}

%% file: figures/overview.tex
\begin{figure*}[!t]
    \centering
    \includegraphics[width=0.98\textwidth]{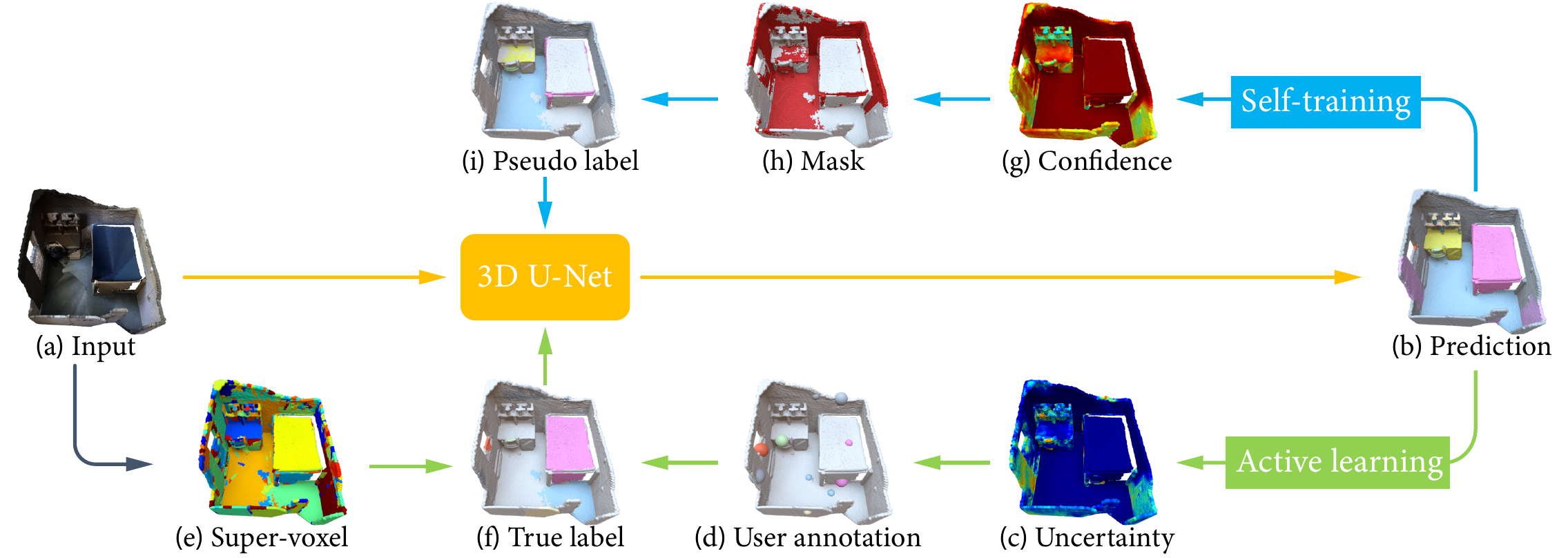}
    \caption{Overview of our method: (a) Given an input scene, we train a neural network to predict (b) a semantic segmentation of the point cloud. Based on the uncertainty of the prediction (c), an active learning method selects a small set of samples which are (d) annotated by a user, where the selected samples are shown as big dots and the samples of previous iterations are shown as small dots. The labels are (f) propagated to the entire point cloud based on (e) an over-segmentation of the points into super-voxels. At the same time, a self-training method selects (i) a set of pseudo labels with (h) the mask determined based on (g) label confidence. The network is then refined based on the (f) true propagated and (i) pseudo labels. This process is then repeated for multiple iterations until a budget of annotations is achieved.}
\label{fig:overview}
\end{figure*}

%% file: figures/tab_comp_scannet.tex

\begin{table}[t!]
\caption{Comparison of our method ActiveST to existing methods and to our fully-supervised baseline on the ScanNet-v2 test set.}
	\begin{minipage}{0.95\columnwidth}
		\begin{center}
			\begin{tabular}{c | c c} \toprule 
				Method & Supervision & mIoU(\%) \\ \midrule 
				PointNet++~\cite{qi17pp} & 100\% & 33.9 \\
				TangentConv~\cite{tatarchenko2018} & 100\% & 43.8 \\
				KPConv~\cite{thomas2019} & 100\% & 68.4 \\
				SubSparseCNN~\cite{graham2018} & 100\% & 72.5 \\
				MinkowskiNet~\cite{choy20194d} & 100\% & 73.6 \\
				Virtual MVFusion~\cite{kundu2020} & 100\%+2D & 74.6 \\ \midrule 
				Our fully-sup baseline & 100\% & 73.6 \\ \midrule 
				1T1C~\cite{liu2021one} & 0.02\% & 69.1 \\
				CSC~\cite{hou2021exploring} & 20 points/scene & 53.1 \\
				PointContrast~\cite{xie2020} & 20 points/scene & 55.0 \\
				1T1C~\cite{liu2021one} & 20 points/scene & 59.4 \\ 
				ActiveST (Ours) & 20 points/scene & \bf{70.3} \\ \bottomrule 
			\end{tabular}
		\end{center}
	\end{minipage}
	\label{tab:comp_scannet}
\end{table}

%% file: figures/comp.tex
\begin{figure*}[!t]
	\centering
	\includegraphics[width=\textwidth]{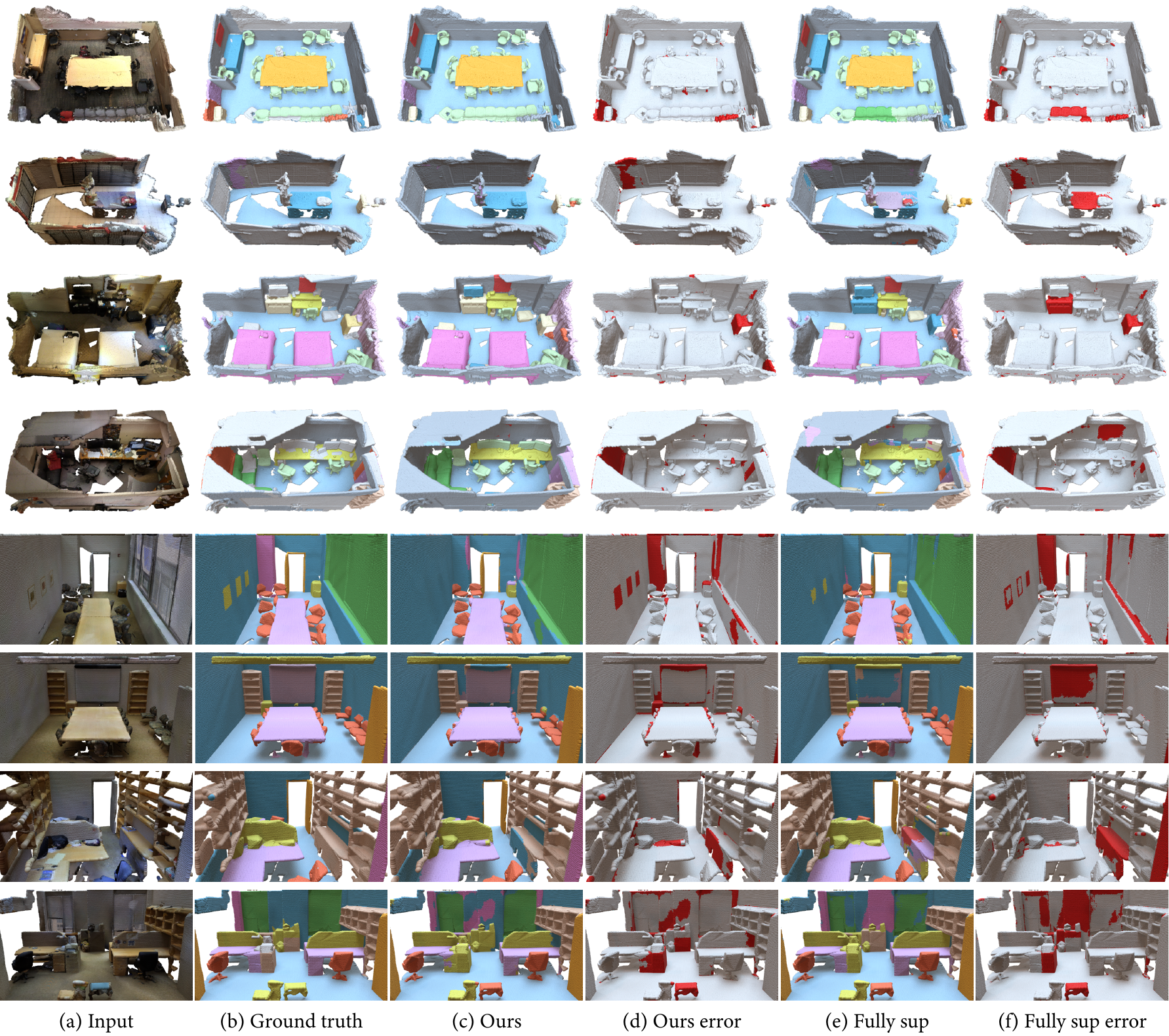}
        \caption{Qualitative segmentation results: the first four rows are from the ScanNet-v2 dataset and the following four rows are from the S3DIS dataset. The input and the ground truth segmentations are presented in (a) and (b). (c) and (e) are the prediction of our method and of the fully-supervised baseline, while (d) and (f) are the corresponding error maps, where red regions indicate incorrect predictions.} 
	\label{fig:comp}
\end{figure*}

%% file: figures/tab_comp_S3DIS.tex

\begin{table}[t!]
	\centering
	\caption{Comparison of our method ActiveST to existing methods and to our fully-supervised baseline on Area-5 of S3DIS.}
	\begin{minipage}{0.95\columnwidth}
		\begin{center}
			\begin{tabular}{ c | c c} \toprule  
				Method & Supervision & mIoU(\%) \\ \midrule 
				PointNet++~\cite{qi17pp} & 100\% & 41.1 \\
				TangentConv~\cite{tatarchenko2018} & 100\% & 52.8 \\
				3D RNN~\cite{ye20183drnn} & 100\% & 53.4 \\
				SuperpointGraph~\cite{landrieu2018} & 100\% & 58.0 \\
				MinkowskiNet~\cite{choy20194d} & 100\% & 66.3 \\ 
				Virtual MVFusion~\cite{kundu2020} & 100\%+2D & 65.4 \\ \midrule  
				Our fully-sup baseline & 100\% & 66.3 \\ \midrule  
				1T3C~\cite{liu2021one} & 0.06\% & 55.3 \\ 
				1T1C~\cite{liu2021one} & 0.02\% & 50.1 \\
				ActiveST (Ours) & 20 points/scene & \bf{56.3} \\ \bottomrule 
			\end{tabular}
		\end{center}
	\end{minipage}
	\label{tab:comp_S3DIS}
\end{table}

%% file: figures/tab_ablation.tex
\begin{table}[t!]
    \caption{Ablation studies of our method conducted with ``20 points/scene'' annotation. ScanNet-v2 is evaluated on the validation set while S3DIS is evaluated on Area 5. ``Voting'' indicates averaging the prediction of all the points inside of the same super-voxel during inference. ``Self-train.'' refers to the self-training approach used to generate the pseudo label set to train the segmentation network. ``Active learn." refers to the active learning method used to select the samples for annotation, which are propagated to constitute a per-point true label set. }
	\begin{tabular}{ccc | cc} \toprule  
		\multicolumn{3}{c |}  {Components} & \multicolumn{2}{c}{mIoU(\%)}  \\ \midrule 
		Voting & Self-train. & Active learn. & ScanNet & S3DIS \\ \midrule  
		  &   &   & 59.2 & 39.5 \\
		\checkmark & &     & 62.3 & 40.5 \\
		\checkmark & \checkmark &   & 67.2 & 47.2 \\
		\checkmark &   & \checkmark & 65.2 & 51.8 \\
		\checkmark & \checkmark & \checkmark & \bf{69.8} & \bf{56.3} \\ \bottomrule  
	\end{tabular}
	\label{tab:ablation}
\end{table}

%% file: figures/sample.tex
\begin{figure*}[!t]
	\centering
	\includegraphics[width=\textwidth]{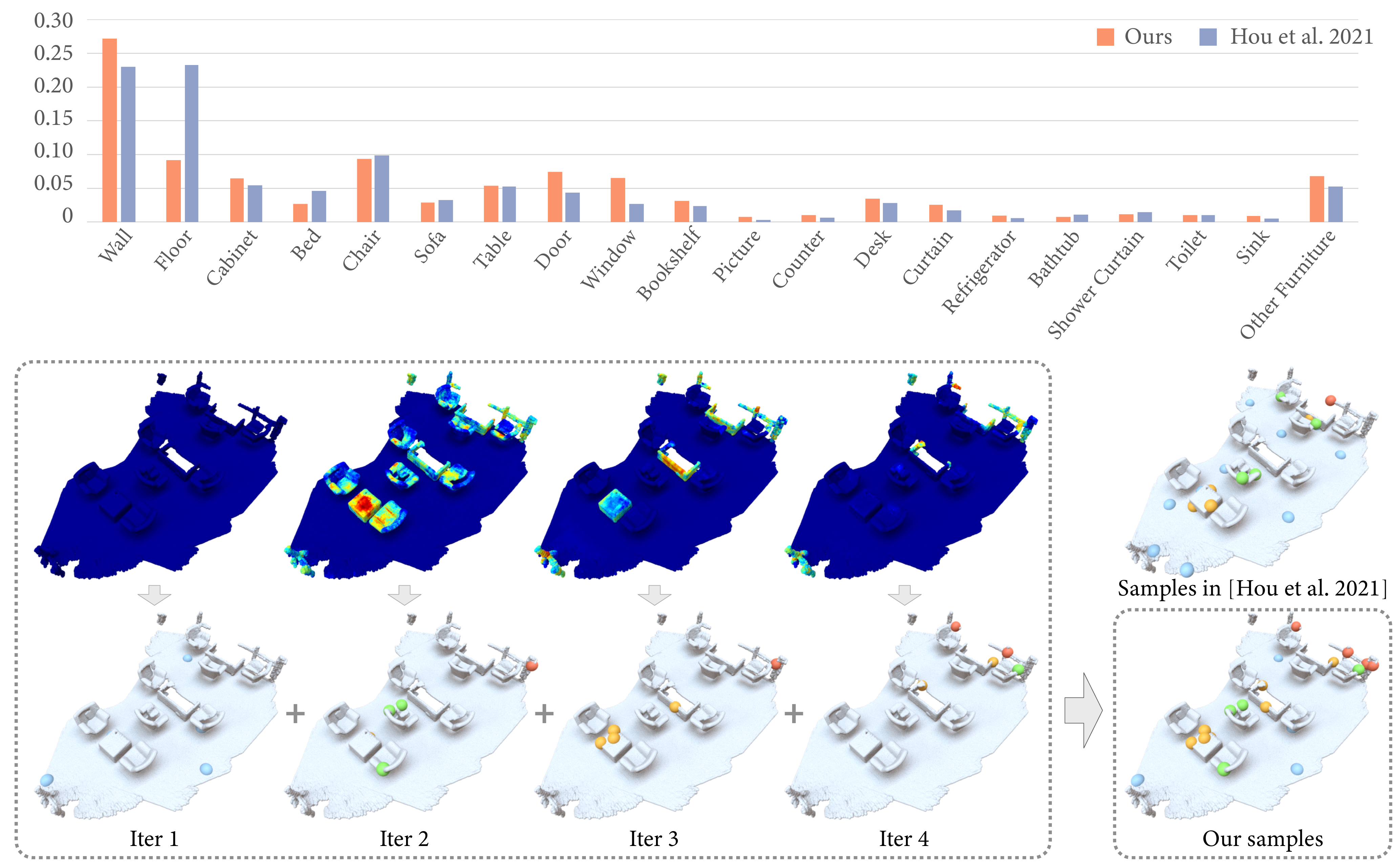}
        \caption{Comparison of samples selected for annotation. Top: statistics on the classes of objects where points were selected with the active learning in our method, compared to~\cite{hou2021exploring}. Bottom: a visual comparison between the samples that the active learning selected based on uncertainty across four iterations and those selected at once based on a pre-trained model~\cite{hou2021exploring} (top-right scene outside the dashed box). }
	\label{fig:sample}
\end{figure*}

%% file: figures/tab_1T1C.tex
\begin{table}[t!]
    \caption{Evaluation of our method in the ``one thing one click'' setting, selecting one point per object during the active learning.}
	\centering
	\begin{minipage}{\columnwidth}
		\begin{center}
            \begin{tabular}{c | c c}  \toprule
        		\multirow{2}{*}{Method} & \multicolumn{2}{c}{mIoU(\%)}  \\ \cline{2-3}
        		 & ScanNet & S3DIS \\ \midrule
				1T1C \cite{liu2021one} & 70.5 & 50.1 \\  \midrule
				Iter 1 (+6 points) & 45.7 & 36.5 \\
				Iter 2 (+6 points) & 62.5 & 45.8 \\
				Iter 3 (+6 points) & 66.9 & 48.0 \\
				Iter 4 (+6 points) & 68.7 & 51.2 \\
				Iter 5 (+6 points) & 69.3 & 53.3 \\ 
        		\multirow{2}{*}{ 
        		\makecell{Iter 6 (+1 point for \\ each remaining object) }} 
        		& \multirow{2}{*}{\bf{71.5}} & \multirow{2}{*}{\bf{54.9}} \\
        		 &  &  \\ \bottomrule
			\end{tabular}
		\end{center}
	\end{minipage}
	\label{tab:1T1C}
\end{table}

%% file: figures/sample_within_object.tex
\begin{figure}[!t]
	\centering
	\includegraphics[width=\linewidth]{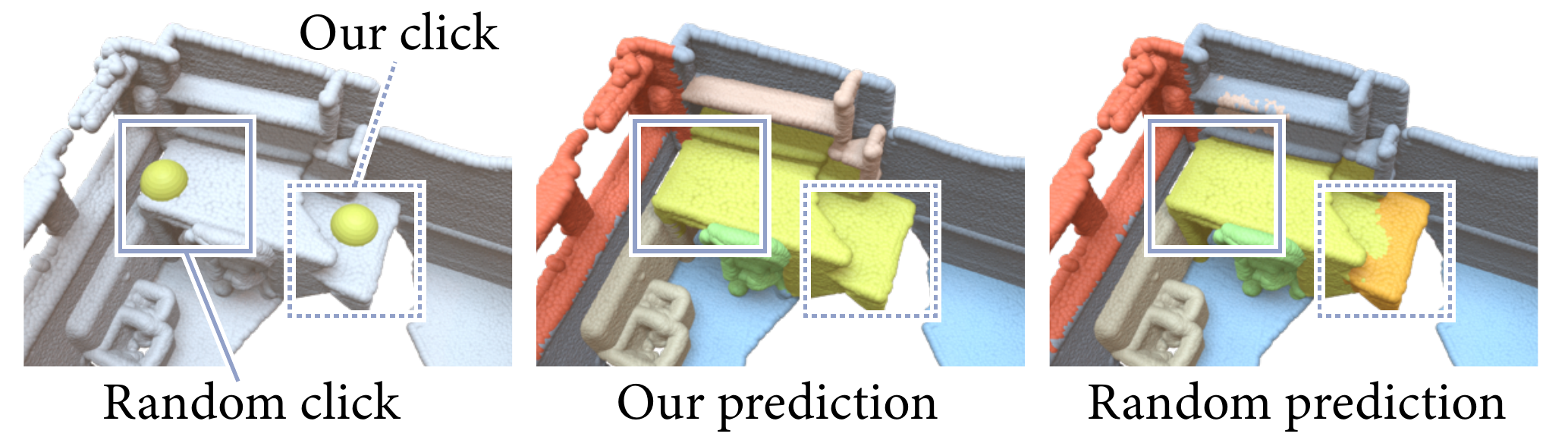}
	\caption{Visual comparison of a sample point selected by our method and the point randomly sampled as in~\cite{liu2021one}, with the corresponding prediction results. 
	We see that our sample is located on a more challenging region of the desk which leads to a more accurate prediction after training. }
	\label{fig:sample_within_object}
\end{figure}

%% file: figures/sample_number.tex
\begin{figure}[!t]
	\centering
	\includegraphics[width=\linewidth]{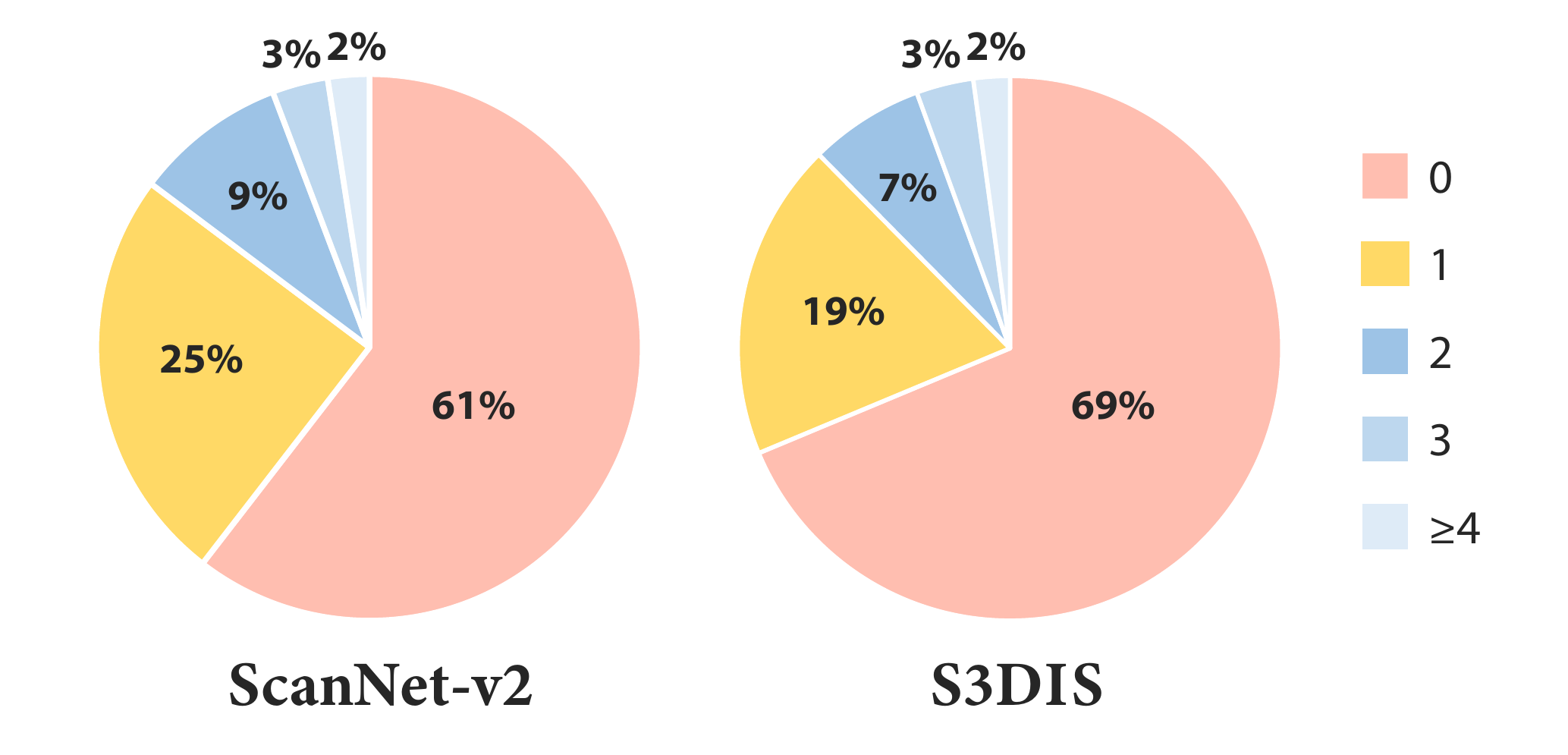}
	\caption{Statistics on the number of points sampled on each object by our method under the 20 points/scene setting. }
	\label{fig:sample_number}
\end{figure}

%% file: figures/network.tex
\begin{figure}[h]
	\centering
    \includegraphics[width=0.9\columnwidth]{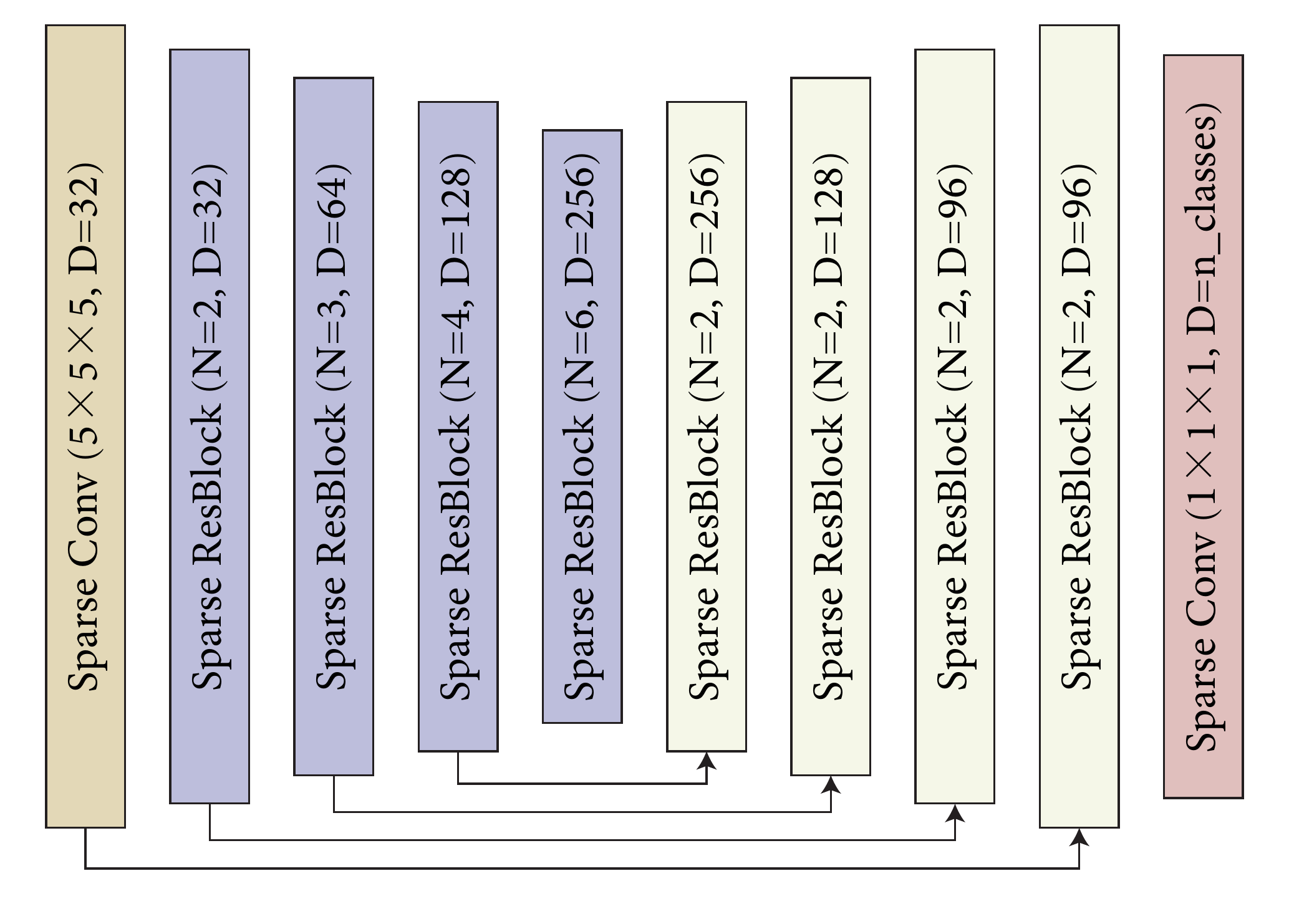}
        \caption{3D U-Net architecture~\cite{choy20194d} used as our backbone.} 
	\label{fig:network}
\end{figure}

%% file: figures/algorithm.tex
\begin{algorithm}[h] 
\renewcommand{\algorithmicrequire}{\textbf{Input:}}
\renewcommand\algorithmicensure {\textbf{Output:} }
\caption{ Training procedure of our ActiveST framework.}
\label{algorithm:activest} 
\begin{algorithmic}[1] 

\REQUIRE Total amount of user annotations $n$, number of iterations $k$, super-voxel partition;

\STATE Random sample $m=n/k$ points to annotate, and propagate the labels in super-voxels to get $T$;

\STATE Train a network $\theta$ using $T$;

\FOR{$i=1$ to $k-1$}  
    \STATE Use $\theta$ to generate pseudo-labels and compute uncertainty and confidence;
    
    \STATE Select $m$ points to annotate according to the uncertainty distribution, and propagate the labels in super-voxels to get $T_{i}$;
    
    \STATE Update true labels $T = T\cup T_{i}$;
    
    \STATE Select pseudo-labels with high confidence on unlabeled points to get $P$;
    
    \STATE Train a new network $\theta_{i}$ using $T$ and $P$ with the softmax cross-entropy loss;
    
    \STATE $\theta \leftarrow \theta_{i}$
\ENDFOR

\ENSURE Segmentation network $\theta$.
\end{algorithmic}
\end{algorithm}

%% file: figures/tab_sup_percategory.tex
\begin{sidewaystable}
	\caption{Per-category performance of our method compared to other weakly supervised methods on the ScanNet-v2 data-efficient benchmark (20 labeled points per scene for training).}
	\label{tab:sup_percategory}
	\addtolength{\tabcolsep}{-4.3pt}
	\begin{center}
		\begin{tabular}{ c | c | c c c c c c c c c c c c c c c c c c c c c} \toprule  
			
			Method  & mIoU & bathtub & bed & booksh. & cabinet & chair & counter & curtain & desk & door & floor & otherfur. & picture & refrigerator & shower. & sink & sofa & table & toilet & wall & window \\ \midrule 
			
			\cite{hou2021exploring} & 53.1 & 65.9	&63.8	&57.8	&41.7	&77.5	&25.4	&53.7	&39.6	&43.9&	93.9	&28.4	&8.3	&41.4	&59.9	&48.8	&69.8	&44.4	&78.5	&74.7	&44.0 \\ 
			\cite{xie2020} & 55.0 & 73.5&	67.6&	60.1&	47.5&	79.4&	28.8	&62.1&	37.8&	43.0&	94.0&	30.3&	8.9&	37.9&	58.0&	53.1&	68.9&	42.2&	85.2&	75.8&	46.8 \\
			\cite{liu2021one} & 59.4 & 75.6&	72.2&	49.4&	54.6&	79.5&	37.1&	72.5&	55.9&	48.8&	95.7&	36.7&	\bf{26.1}&	54.7&	57.5&	22.5&	67.1&	54.3&	90.4&	82.6&	55.7 \\ \midrule
			Ours & \bf{70.3} & \bf{97.7}&	\bf{77.6}&	\bf{65.7}&	\bf{70.7}&	\bf{87.4}&	\bf{54.1}&	\bf{74.4}&	\bf{60.5}&	\bf{61.0}&	\bf{96.8}&	\bf{44.2}&	12.6&	\bf{70.5}&	\bf{78.5}&	\bf{74.2}&	\bf{79.1}&	\bf{58.6}&	\bf{94.0}&	\bf{83.9}&	\bf{64.5} \\ \bottomrule 
		\end{tabular}
	\end{center}
\end{sidewaystable}

%% file: figures/sup_comp_active_benchmark.tex
\begin{figure*}[t!]
	\centering
    \includegraphics[width=0.81\textwidth]{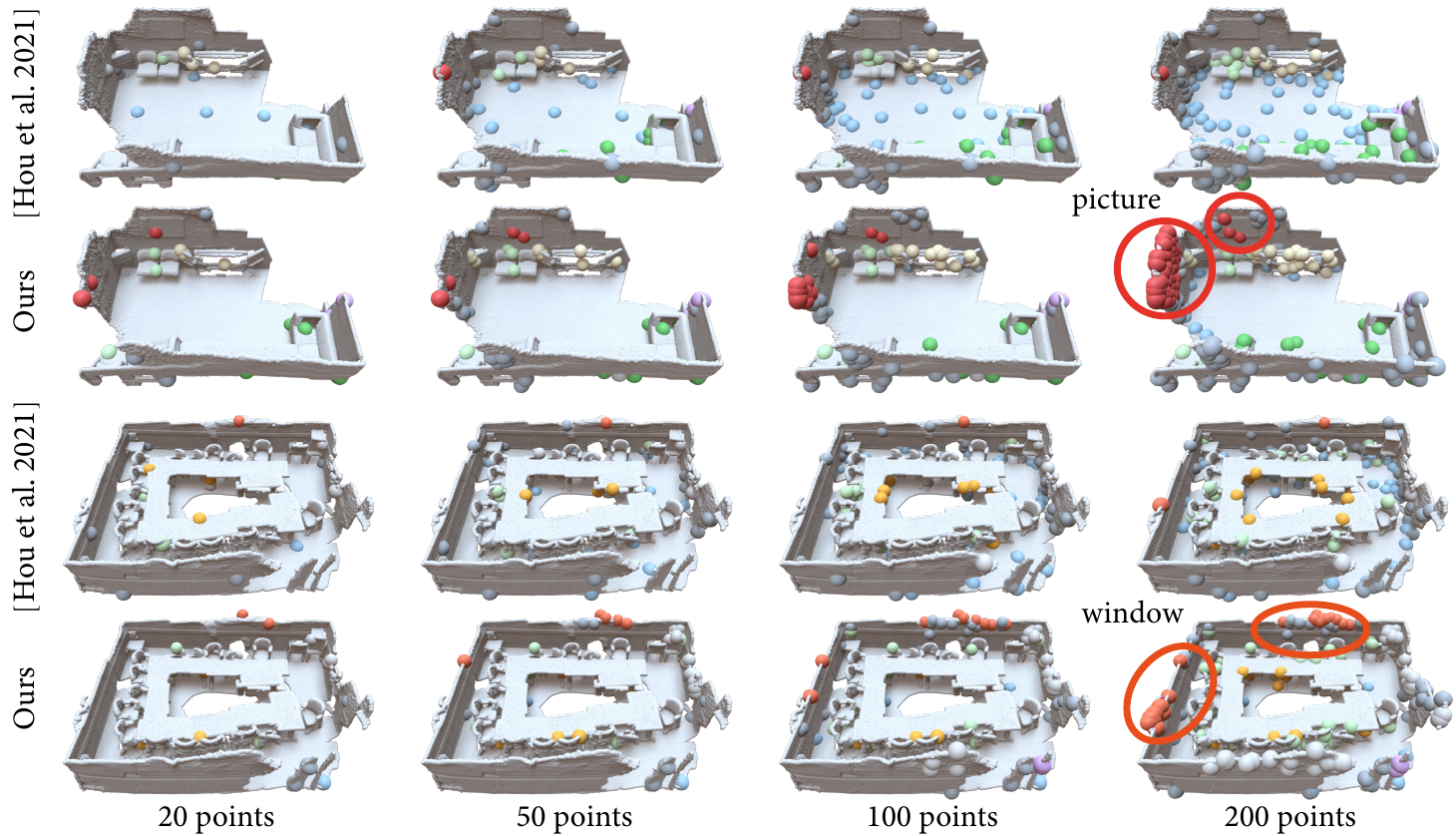}
        \caption{Comparison of samples selected by our method and the method of~\cite{hou2021exploring} with different annotation budget.}
	\label{fig:sup_comp_active_benchmark}
\end{figure*}

%% file: figures/tab_sup_scannet.tex
\begin{table}[t]
	\centering
	\addtolength{\tabcolsep}{-1.5pt}
	\caption{Comparison  of  our  method (ActiveST) to other methods under different limited point annotations, where ``pts'' means points/scene. We report mIoU(\%) on Scanet-v2 test set. }
	\begin{minipage}{\columnwidth}
		\begin{center}
			\begin{tabular}{ c | c c c c} \toprule  
				Method & 20 pts & 50 pts & 100 pts & 200 pts \\ \midrule 
				CSC~\cite{hou2021exploring} & 53.1 & 61.2 & 64.4  & 66.5 \\
				PointContrast~\cite{xie2020} & 55.0 & 61.4 & 63.5 & 65.3 \\
				1T1C~\cite{liu2021one} & 59.4 & 64.2 & 67.0 & 69.4 \\
				ActiveST (Ours) & \bf{70.3} & \bf{72.5} & \bf{73.5} & \bf{74.8}  \\ \bottomrule 
			\end{tabular}
		\end{center}
	\end{minipage}
	\label{tab:sup_scannet}
\end{table}

%% file: figures/tab_sup_s3dis.tex
\begin{table}[t]
	\centering
	\caption{Comparison of our method (ActiveST) to our fully-supervised baseline under different limited point annotations. We report mIoU(\%) on Area-5 of S3DIS.}
	\begin{minipage}{\columnwidth}
		\begin{center}
			\begin{tabular}{ c | c c} \toprule  
				Method & Supervision & mIoU(\%) \\ \midrule 
				ActiveST (Ours) & 20 pts & 56.3 \\
				ActiveST (Ours) & 50 pts & 57.8 \\  
				ActiveST (Ours) & 100 pts & 59.5 \\
				ActiveST (Ours) & 200 pts & 62.1 \\ \midrule
				
				Our fully-sup baseline & 100\% & 66.3 \\ \bottomrule
			\end{tabular}
		\end{center}
	\end{minipage}
	\label{tab:sup_s3dis}
\end{table}

%% file: cvmpaper_finalcopy.bbl
\begin{thebibliography}{9}
	
	\bibitem{armeni2017joint} Armeni, I., Sax, S., Zamir, A. R.,  Savarese, S. Joint 2d-3d-semantic data for indoor scene understanding. arXiv preprint arXiv:1702.01105, 2017.
	
	\bibitem{cheng2021sspc} Cheng, M., Hui, L., Xie, J.,  Yang, J. SSPC-Net: Semi-supervised semantic 3D point cloud segmentation network. arXiv preprint arXiv:2104.07861, 2021.
	
	
	\bibitem{choy20194d} Choy, C., Gwak, J.,  Savarese, S. 4d spatio-temporal convnets: Minkowski convolutional neural networks. In Proceedings of the IEEE/CVF Conference on Computer Vision and Pattern Recognition (CVPR), 2019, 3075-3084. 
	
	\bibitem{dai2017scannet} Dai, A., Chang, A. X., Savva, M., Halber, M., Funkhouser, T.,  Nießner, M. Scannet: Richly-annotated 3d reconstructions of indoor scenes. In Proceedings of the IEEE/CVF Conference on Computer Vision and Pattern Recognition (CVPR), 2017, 5828-5839.
	
	\bibitem{dai2018} Dai, A.,  Nießner, M. 3dmv: Joint 3d-multi-view prediction for 3d semantic scene segmentation. In Proceedings of the European Conference on Computer Vision (ECCV), 2018, 452-468.
	
	\bibitem{gadelha2020} Gadelha, M., RoyChowdhury, A., Sharma, G., Kalogerakis, E., Cao, L., Learned-Miller, E., Maji, S. Label-efficient learning on point clouds using approximate convex decompositions. In Proceedings of the European Conference on Computer Vision (ECCV), 2020, 473-491.
	
	\bibitem{graham2018} Graham, B., Engelcke, M.,  Van Der Maaten, L. 3d semantic segmentation with submanifold sparse convolutional networks. In Proceedings of the IEEE/CVF Conference on Computer Vision and Pattern Recognition (CVPR), 2018, 9224-9232.
	
	\bibitem{han2020} Han, L., Zheng, T., Xu, L.,  Fang, L. Occuseg: Occupancy-aware 3d instance segmentation. In: Proceedings of the IEEE/CVF Conference on Computer Vision and Pattern Recognition (CVPR), 2020, 2940-2949.
	
	\bibitem{hou2021exploring} Hou, J., Graham, B., Nießner, M.,  Xie, S. Exploring data-efficient 3d scene understanding with contrastive scene contexts. In Proceedings of the IEEE/CVF Conference on Computer Vision and Pattern Recognition (CVPR), 2021, 15587-15597.
	
	\bibitem{hu18semantic} Hu, R., Wen, C., Van Kaick, O., Chen, L., Lin, D., Cohen-Or, D.,  Huang, H. Semantic object reconstruction via casual handheld scanning. ACM Transactions on Graphics (TOG) Vol.37, No.6, 1-12, 2018.
	
	\bibitem{jiang2021} Jiang, L., Shi, S., Tian, Z., Lai, X., Liu, S., Fu, C. W.,  Jia, J. Guided Point Contrastive Learning for Semi-supervised Point Cloud Semantic Segmentation. In Proceedings of the IEEE/CVF International Conference on Computer Vision (ICCV), 2021, 6423-6432.
	
	\bibitem{komarichev2019} Komarichev, A., Zhong, Z.,  Hua, J. A-cnn: Annularly convolutional neural networks on point clouds. In Proceedings of the IEEE/CVF Conference on Computer Vision and Pattern Recognition (CVPR), 2019, 7421-7430.
	
	\bibitem{kundu2020} Kundu, A., Yin, X., Fathi, A., Ross, D., Brewington, B., Funkhouser, T.,  Pantofaru, C. Virtual multi-view fusion for 3d semantic segmentation. In Proceedings of the European Conference on Computer Vision (ECCV), 2020, 518-535.
	
	\bibitem{landrieu2018} Landrieu, L.,  Simonovsky, M. Large-scale point cloud semantic segmentation with superpoint graphs. In Proceedings of the IEEE/CVF Conference on Computer Vision and Pattern Recognition (CVPR), 2018, 4558-4567.
	
	\bibitem{li2018} Li, Y., Bu, R., Sun, M., Wu, W., Di, X.,  Chen, B. Pointcnn: Convolution on x-transformed points. 
	Advances in neural information processing systems (NIPS), 2018, 31.
	
	\bibitem{liu2019} Liu, Z., Tang, H., Lin, Y.,  Han, S. Point-voxel cnn for efficient 3d deep learning. Advances in neural information processing systems (NIPS), 2019, 32.
	
	\bibitem{liu2021one} Liu, Z., Qi, X.,  Fu, C. W. One Thing One Click: A Self-Training Approach for Weakly Supervised 3D Semantic Segmentation. In Proceedings of the IEEE/CVF Conference on Computer Vision and Pattern Recognition (CVPR), 2021, 1726-1736.
	
	\bibitem{paszke2019pytorch} Paszke, A., Gross, S., Massa, F., Lerer, A., Bradbury, J., Chanan, G., ...  Chintala, S. Pytorch: An imperative style, high-performance deep learning library. Advances in neural information processing systems (NIPS), 2019, 32.
	
	\bibitem{qi17pp} Qi, C. R., Yi, L., Su, H.,  Guibas, L. J. Pointnet++: Deep hierarchical feature learning on point sets in a metric space. Advances in neural information processing systems (NIPS), 2017, 30.
	
	\bibitem{qi17} Qi, C. R., Su, H., Mo, K.,  Guibas, L. J. Pointnet: Deep learning on point sets for 3d classification and segmentation. In Proceedings of the IEEE/CVF Conference on Computer Vision and Pattern Recognition (CVPR), 2017, 652-660.
	
	\bibitem{rizve2021defense} Rizve, M. N., Duarte, K., Rawat, Y. S.,  Shah, M. In defense of pseudo-labeling: An uncertainty-aware pseudo-label selection framework for semi-supervised learning. arXiv preprint arXiv:2101.06329, 2021.
	
	\bibitem{shi2021label} Shi, X., Xu, X., Chen, K., Cai, L., Foo, C. S.,  Jia, K. Label-efficient point cloud semantic segmentation: An active learning approach. arXiv preprint arXiv:2101.06931, 2021.
	
	\bibitem{su2018} Su, H., Jampani, V., Sun, D., Maji, S., Kalogerakis, E., Yang, M. H.,  Kautz, J. Splatnet: Sparse lattice networks for point cloud processing. In Proceedings of the IEEE/CVF Conference on Computer Vision and Pattern Recognition (CVPR), 2018, 2530-2539.
	
	\bibitem{tatarchenko2018} Tatarchenko, M., Park, J., Koltun, V.,  Zhou, Q. Y. Tangent convolutions for dense prediction in 3d. In Proceedings of the IEEE/CVF Conference on Computer Vision and Pattern Recognition (CVPR), 2018, 3887-3896. 
	
	\bibitem{thomas2019} Thomas, H., Qi, C. R., Deschaud, J. E., Marcotegui, B., Goulette, F.,  Guibas, L. J. Kpconv: Flexible and deformable convolution for point clouds. In Proceedings of the IEEE/CVF Conference on Computer Vision and Pattern Recognition (CVPR), 2019, 6411-6420.
	
	\bibitem{wei2020} Wei, J., Lin, G., Yap, K. H., Hung, T. Y.,  Xie, L. Multi-path region mining for weakly supervised 3D semantic segmentation on point clouds. In Proceedings of the IEEE/CVF Conference on Computer Vision and Pattern Recognition (CVPR), 2020, 4384-4393.
	
	\bibitem{wu2021} Wu, T. H., Liu, Y. C., Huang, Y. K., Lee, H. Y., Su, H. T., Huang, P. C.,  Hsu, W. H. ReDAL: Region-based and Diversity-aware Active Learning for Point Cloud Semantic Segmentation. In Proceedings of the IEEE/CVF International Conference on Computer Vision (ICCV), 2021, 15510-15519.
	
	\bibitem{wu2015} Wu, Z., Song, S., Khosla, A., Yu, F., Zhang, L., Tang, X.,  Xiao, J. 3d shapenets: A deep representation for volumetric shapes. In Proceedings of the IEEE/CVF Conference on Computer Vision and Pattern Recognition (CVPR), 2015, 1912-1920.
	
	\bibitem{wu2019} Wu, W., Qi, Z.,  Fuxin, L. Pointconv: Deep convolutional networks on 3d point clouds. In Proceedings of the IEEE/CVF Conference on Computer Vision and Pattern Recognition (CVPR), 2019, 9621-9630.
	
	\bibitem{xu2020} Xu, X.,  Lee, G. H. Weakly supervised semantic point cloud segmentation: Towards 10x fewer labels. In Proceedings of the IEEE/CVF Conference on Computer Vision and Pattern Recognition (CVPR), 2020, 13706-13715.
	
	\bibitem{xie2020} Xie, S., Gu, J., Guo, D., Qi, C. R., Guibas, L.,  Litany, O. Pointcontrast: Unsupervised pre-training for 3d point cloud understanding. In European conference on computer vision (ECCV), 2020, 574-591.
	
	\bibitem{yi2016} Yi, L., Kim, V. G., Ceylan, D., Shen, I. C., Yan, M., Su, H.,  Guibas, L. A scalable active framework for region annotation in 3d shape collections. ACM Transactions on Graphics (ToG) Vol.35, No.6, 1-12, 2016.
	
	\bibitem{ye20183drnn} Ye, X., Li, J., Huang, H., Du, L.,  Zhang, X. 3d recurrent neural networks with context fusion for point cloud semantic segmentation. In European conference on computer vision (ECCV), 2018, 403-417.
	
	\bibitem{zhang2021} Zhang, Z., Girdhar, R., Joulin, A.,  Misra, I. Self-supervised pretraining of 3d features on any point-cloud. IEEE/CVF International Conference on Computer Vision (ICCV), 2021, 10252-10263.
	
	\bibitem{guo2021pct} Guo, M. H., Cai, J. X., Liu, Z. N., Mu, T. J., Martin, R. R., Hu, S. M. Pct: Point cloud transformer. Computational Visual Media, 2021, 7(2): 187-199.
	
	\bibitem{zhang2020fusion} Zhang, J., Zhu, C., Zheng, L., Xu, K. Fusion-aware point convolution for online semantic 3d scene segmentation. In Proceedings of the IEEE/CVF Conference on Computer Vision and Pattern Recognition (CVPR), 2020, 4534-4543.
	
	\bibitem{huang2021supervoxel} Huang, S. S., Ma, Z. Y., Mu, T. J., Fu, H., Hu, S. M. Supervoxel convolution for online 3d semantic segmentation. ACM Transactions on Graphics (ToG) Vol.40, No.3, 1-15, 2021.
	
	\bibitem{lin2018toward} Lin, Y., Wang, C., Zhai, D., Li, W., Li, J. Toward better boundary preserved supervoxel segmentation for 3D point clouds. ISPRS journal of photogrammetry and remote sensing, 143, 39-47, 2018.
	
	\bibitem{peng2020semantic} Peng, Haotian and Zhou, Bin and Yin, Liyuan and Guo, Kan and Zhao, Qinping. Semantic part segmentation of single-view point cloud. Sci China Inf Sci, 2020, 63(12): 224101.
	
\end{thebibliography}
